\documentclass[runningheads]{llncs}

\pdfoutput=1

\usepackage{graphicx}
\usepackage{amsmath,amssymb} 
\usepackage{color}

\usepackage{epsfig}
\usepackage{graphicx}
\usepackage{amsmath}
\usepackage{amssymb}

\usepackage{authblk}

\usepackage{algorithm}
\usepackage{algorithmic}
\usepackage{amsmath}
\usepackage{enumerate}
\usepackage{placeins}
\usepackage[symbol]{footmisc}

\usepackage[width=122mm,left=12mm,paperwidth=146mm,height=193mm,top=12mm,paperheight=217mm]{geometry}

\newcommand{\fig}[1]{Figure~\ref{fig:#1}}
\newcommand{\sect}[1]{Section~\ref{sect:#1}}
\newcommand{\tab}[1]{Table~\ref{tab:#1}}

\newcommand{\eq}[1]{(\ref{eq:#1})}

\renewcommand{\lastandname}{\unskip, }

\begin{document}

\pagestyle{headings}
\mainmatter
\def\ECCV18SubNumber{1740}  

\title{Revisiting the Inverted Indices for Billion-Scale Approximate Nearest Neighbors} 

\titlerunning{Revisiting the Inverted Indices for Billion-Scale ANN}
\authorrunning{Baranchuk D., Babenko A., Malkov Y.}

\author{Dmitry Baranchuk\inst{1,2} \lastandname Artem Babenko\inst{1,3} \lastandname Yury Malkov\inst{4}}
\institute{Yandex \and Lomonosov Moscow State University \and National Research University Higher School of Economics \and The Institute of Applied Physics of the Russian Academy of Sciences}

\maketitle

\begin{abstract}
This work addresses the problem of billion-scale nearest neighbor search. The state-of-the-art retrieval systems for billion-scale databases are currently based on the inverted multi-index, the recently proposed generalization of the inverted index structure. The multi-index provides a very fine-grained partition of the feature space that allows extracting concise and accurate short-lists of candidates for the search queries.

In this paper, we argue that the potential of the simple inverted index was not fully exploited in previous works and advocate its usage both for the highly-entangled deep descriptors and relatively disentangled SIFT descriptors. We introduce a new retrieval system that is based on the inverted index and outperforms the multi-index by a large margin for the same memory consumption and construction complexity.
For example, our system achieves the state-of-the-art recall rates several times faster on the dataset of one billion deep descriptors compared to the efficient implementation of the inverted multi-index from the FAISS library. 

\end{abstract}

\section{Introduction}

The last decade efficient billion-scale nearest neighbor search has become a significant research problem\cite{Jegou11b,Babenko12,OpqTr,Kalantidis14,BabenkoCVPR16,FAISS}, inspired by the needs of modern computer vision applications, e.g. large-scale visual search\cite{Philbin07}, low-shot classification\cite{Douze_2018_CVPR} and face recognition\cite{Wang_2017_TPAMI}. In particular, since the number of images on the Internet grows enormously fast, the multimedia retrieval systems need scalable and efficient search algorithms to respond queries to the databases of billions of items in several milliseconds.

All the existing billion-scale systems avoid the infeasible exhaustive search via restricting the part of the database that is considered for a query. This restriction is performed with the help of an \textit{indexing structure}. The indexing structures partition the feature space into a large number of disjoint regions, and the search process inspects only the points from the regions that are the closest to the particular query. The inspected points are organized in \textit{short-lists} of candidates and the search systems calculate the distances between the query and all the candidates exhaustively. In scenarios, when the database does not fit in RAM, the compressed representations of the database points are used. The compressed representations are typically obtained with product quantization\cite{Jegou11a} that allows to compute the distances between the query and compressed points efficiently. The step of the distances calculation has a complexity that is linear in the number of candidates hence the short-lists provided by indexing structures should be concise.

The first indexing structure that was able to operate on the billion-scale datasets was introduced in \cite{Jegou11b}. It was based on the inverted index structure that splits the feature space into Voronoi regions for a set of K-Means centroids, learned on the dataset. This system was shown to achieve reasonable recall rates in several tens of milliseconds.

Later a generalization of the inverted index structure was proposed in \cite{Babenko12}. This work introduced the inverted multi-index (IMI) that decomposes the feature space into several orthogonal subspaces and partitions each subspace into Voronoi regions independently. Then the Cartesian product of regions in each subspace formes the implicit partition of the whole feature space. Due to a huge number of regions, the IMI space partition is very fine-grained, and each region contains only a few data points. Therefore, IMI forms accurate and concise candidate lists while being memory and runtime efficient.

However, the structured nature of the regions in the IMI partition also has a negative impact on the final retrieval performance. In particular, it was shown in \cite{BabenkoCVPR16} that the majority of IMI regions contain no points and the effective number of regions is much smaller than the theoretical one. For certain data distributions, this results in the fact that the search process spends much time visiting empty regions that produce no candidates. In fact, the reason for this deficiency is that the IMI learns K-Means codebooks independently for different subspaces while the distributions of the corresponding data subvectors are not statistically independent in practice. In particular, there are significant correlations between different subspaces of CNN-produced descriptors that are most relevant these days. 
In this paper, we argue that the previous works underestimate the simple inverted index structure and advocate its use for all data types. The contributions of our paper include:

\begin{enumerate}
    \item We demonstrate that the performance of the inverted index could be substantially boosted via using larger codebooks, while the multi-index design does not allow such a boost. 
    \item We introduce a memory-efficient \textit{grouping} procedure for database points that boosts retrieval performance even further. 
    \item We provide an optimized implementation of our system for billion-scale search in the compressed domain to support the following research on this problem. As we show, the proposed system achieves the state-of-the-art recall rates up to several times faster, compared to the advanced IMI implementation from the FAISS library\cite{FAISS} for the same memory consumption. The C++ implementation of our system is publicly available online\footnote[1]{https://github.com/dbaranchuk/ivf-hnsw}.
\end{enumerate}

The paper is structured as follows. We review related works on billion-scale indexing in Section 2. Section 3 describes a new system based on the inverted index. The experiments demonstrating the advantage of our system are detailed in Section 4. Finally, Section 5 concludes the paper.

\section{Related work}

In this section we briefly review the previous methods that are related to our approach. Also here we introduce notation for the following sections. 

\textbf{Product quantization (PQ)} is a lossy compression method for high-dimensional vectors~\cite{Jegou11a}. Typically, PQ is used in scenarios when the large-scale datasets do not fit into the main memory. In a nutshell, PQ encodes each vector $x \in \mathbf{R}^D$ by a concatenation of $M$ codewords from $M$ $\frac{D}{M}$\nobreakdash-dimensional codebooks $R_1,\ldots,R_M$. Each codebook typically contains $256$ codewords $R_m = \{r_1^m,\dots,r_{256}^m\} \subset \mathbf{R}^D$ so that the codeword id could fit into one byte. In other words, PQ decomposes a vector $x$ into $M$ separate subvectors $\left[x_1,\dots,x_M\right]$ and applies vector quantization (VQ) to each subvector $x_m$, while using a separate codebook $R_m$. Then the $M$-byte code for the vector $x$ is a tuple of codewords indices $[i_1,\ldots,i_M]$ and the effective approximation is $x \approx [r_{i_1}^1,\ldots, r_{i_M}^M]$.
As a nice property, PQ allows efficient computation of Euclidean distances between the uncompressed query and the large number of compressed vectors. The computation is performed via the ADC procedure \cite{Jegou11a} using lookup tables:
\begin{gather}
\label{eq:adc}
\|q - x\|^2 \approx \|q - [r_{i_1}^1,\ldots, r_{i_M}^M]\|^2 = \sum\limits_{m=1}^{M}{\|q_m - r_{i_m}^m\|}^2
\end{gather}
where $q_m$ is the $m$th subvector of a query $q$. This sum can be calculated in $M$ additions and lookups given that distances from query subvectors to codewords are precomputed and stored in lookup tables. Thanks to both high compression quality and computational efficiency PQ-based methods are currently the top choice for compact representations of large datasets. PQ gave rise to active research on high-dimensional vectors compression in computer vision and machine learning community\cite{Ge13,Norouzi13,BabenkoCVPR14,BabenkoCVPR15,CQ,SCQ,MartinezECCV16,Polysemous,JegouECCV16}.

\textbf{IVFADC \cite{Jegou11b}} is one of the first retrieval systems capable of dealing with billion-scale datasets efficiently. IVFADC uses the inverted index~\cite{Sivic03} to avoid exhaustive search and Product Quantization for database compression. The inverted index splits the feature space into $K$ regions that are the Voronoi cells of the codebook $C = \{c_1,\dots,c_K\}$. The codebook is typically obtained via standard $K$-means clustering. Then IVFADC encodes the displacements of each point from the centroid of a region it belongs to. The encoding is performed via Product Quantization with global codebooks shared by all regions.

\textbf{The Inverted Multi-Index and Multi-D-ADC.} The inverted multi-index (IMI)~\cite{Babenko12} generalizes the inverted index and is currently the state-of-the-art indexing approach for high-dimensional spaces and huge datasets. Instead of using the full-dimensional codebook, the IMI splits the feature space into several orthogonal subspaces (usually, two subspaces are considered) and constructs a separate codebook for each subspace. Thus, the inverted multi-index has two $\frac{D}{2}$-dimensional codebooks for different halves of the vector, each with $K$ subspace centroids. The feature space partition then is produced as a Cartesian product of the corresponding subspace partitions. Thus for two subspaces the inverted multi-index effectively produces $K^2$ regions. Even for moderate values of $K$ that is much bigger than the number of regions within the IVFADC system or other systems using inverted indices. Due to a very large number of regions only a small fraction of the dataset should be visited to reach the correct nearest neighbor. \cite{Babenko12} also describes the \textit{multi-sequence} procedure that produces the sequence of regions that are the closest to the particular query. For dataset compression, \cite{Babenko12} also uses Product Quantization with codebooks shared across all cells to encode the displacements of the vectors from region centroids. The described retrieval system is referred to as \emph{Multi-D-ADC}.

The performance of indexing in the Multi-D-ADC scheme can be further improved by using the global data rotation that minimizes correlations between subspaces\cite{OpqTr}. Another improvement\cite{Kalantidis14} introduces the Multi-LOPQ system that uses local PQ codebooks for displacements compression with the IMI structure. 

Several other works consider the problem of the memory-efficient billion-scale search. \cite{BabenkoCVPR16} proposes the modification of the inverted multi-index that uses two non-orthogonal codebooks to produce region centroids. \cite{SCQ} proposes to use Composite Quantization\cite{CQ} instead of Product Quantization to produce the partition centroids. While these modifications were shown to achieve higher recall rates compared to the original multi-index, their typical runtimes are about ten milliseconds that could be prohibitively slow in practical scenarios. Several works investigate efficient GPU implementations for billion-scale search\cite{FAISS,Wieschollek_2016_CVPR}.
In this paper, we focus on the niche of the CPU methods that operate with runtimes about one millisecond.

\section{Inverted Index Revisited}
\label{sect:methods}
In this section we first compare the inverted index to the IMI. In particular, we show that the simple increase of the codebook size could substantially improve the indexing quality for the inverted index while being almost useless for the IMI. Second, we introduce a modification for the inverted index that could be used to boost the indexing performance even further without efficiency drop.

\subsection{Index vs Multi-Index}

We compare the main properties of the inverted index and the IMI in the \tab{ivf_imi}. The top part of the table lists the features that make the IMI the state-of-the-art indexing structure these days: precise candidate lists, fast indexing and query assignment due to small codebook sizes (typically $K$ does not exceed $2^{14}$ for billion-sized databases).

\begin{table*}
\centering
\begin{tabular}{|c|c|c|}
\hline
{\bf Structure} & {\bf Inverted Index} & {\bf Inverted Multi-Index}\\
\hline
Candidate lists quality & Medium & \textbf{High} \\
\hline
Query assignment \& indexing cost & Medium & \textbf{Low} \\
\hline
\hline
\begin{tabular}{@{}c@{}}Number of random memory \\ accesses during search\end{tabular}  & \textbf{Small} & Large \\

\hline
Performance increase from large $K$ & \textbf{High} & Small \\
\hline
Memory consumption scalability & $\mathbf{O(K)}$ & $O(K^2)$\\
\hline
\end{tabular}
\label{tab:ivf_imi}
\caption{Comparison of the main properties of the inverted index and the IMI. $K$ denotes the codebook sizes in both systems. The IMI provides more precise candidate lists and has low indexing and query assignment costs due to smaller codebook sizes. On the other hand, the inverted index requires a smaller number of expensive random memory accesses when searching, and could benefit from large codebooks, while the IMI performance saturates with $K$ about $2^{14}$. Moreover, the increase of $K$ is memory-inefficient in the IMI as its additional memory consumption scales quadratically.}
\end{table*}
Nevertheless, the fine-grained partition in the multi-index imposes several limitations that are summarized in the bottom part of the \tab{ivf_imi}. First, the IMI has to visit much more partition regions compared to the inverted index to accumulate the reasonable number of candidates. Skipping to the next region requires a random memory access operation that is more expensive compared to the sequential PQ-distance computation, especially for short code lengths. A large number of random access operations slows down the search, especially when large number of candidates is needed.

Another property that favors the inverted index is the possibility to increase its codebook size $K$. To the best of our knowledge, the largest codebook sizes used in the index vs multi-index comparison were $2^{17}$ and $2^{14}$ respectively\cite{BabenkoCVPR16}. We argue that the multi-index performance is closer to saturation w.r.t $K$ compared to the inverted index, and the usage of $K > 2^{14}$ would not result in substantially better feature space partition. On the other hand, in the inverted index one could use much larger codebooks compared to $K=2^{17}$ without saturation in the space partition quality. To support this claim, we compare the distances from the datapoints to the closest centroids for the inverted index and the IMI with different $K$ values for the DEEP1B dataset\cite{BabenkoCVPR16} in \tab{large_K}. The smaller distances typically indicate that the centroids represent the actual data distribution better. \tab{large_K} demonstrates that the increase of $K$ in the multi-index results in the much smaller decrease of the closest distances compared to the inverted index. E.g. the 16-fold increase of $K$ from $2^{18}$ to $2^{22}$ in the inverted index results in $18\%$ drop in the average distance. On the other hand, the 16-fold increase of regions number in the IMI partition (that corresponds to the fourfold increase in $K$ from $2^{13}$ to $2^{15}$) results only in $11\%$ drop. We also compare amounts of additional memory consumption required by both systems with different $K$ values to demonstrate that the IMI is memory-inefficient for large codebooks. E.g. for $K=2^{15}$ the inverted multi-index requires about four additional bytes per point for one billion database, that is non-negligible, especially for short code lengths. The reason for the quadratic scalability is that the IMI has to maintain $K^2$ inverted lists to represent the feature space partition.

\begin{table*}
\centering
\renewcommand\arraystretch{0.9}
\begin{tabular}{|c|c|c|c|c|c|}
\hline
\multicolumn{3}{|c|}{Inverted Index} & 
\multicolumn{3}{|c|}{Inverted Multi-Index}\\
\hline
K & Average distance & Memory & K & Average distance & Memory\\
\hline
$2^{18}$ & 0.315 & 97Mb & $2^{13}$ & 0.345 & 256Mb\\
\hline
$2^{20}$ & 0.282 & 388Mb & $2^{14}$ & 0.321 & 1024Mb\\
\hline
$2^{22}$ & 0.259 & 1552Mb & $2^{15}$ & 0.305 & 4096Mb\\
\hline
\end{tabular}
\caption{The indexing quality and the amount of additional memory consumption for the inverted index and the IMI with different codebook sizes on the DEEP1B dataset. The indexing quality is evaluated by the average distance from the datapoints to the closest region centroid. The IMI indexing quality does not benefit from $K>2^{14}$ while the required memory grows quadratically.}
\label{tab:large_K}
\end{table*}

The numbers from \tab{large_K} encourage to use the inverted index with larger codebook instead of the IMI, despite the smaller number of the partition regions. The only practical reason, preventing their usage, is the expensive procedure of query assignment that takes $O(K)$ operations. But in the experimental section below we demonstrate that due to the recent progress in the million-scale ANN-search one can use the approximate search of high accuracy for query assignment. We show that the usage of the approximate search does not result in the search performance drop and the overall scheme of the inverted index with approximate query assignment outperforms the state-of-the-art IMI implementation.

\subsection{Grouping and pruning}
\label{sect:grouping}
Now we describe a technique that is especially useful for the IVFADC scheme for compressed domain search. In general, we propose a procedure that organizes the points in each region into several groups such that the points in nearby locations belong to the same group. In other words, we want to split each inverted index region into a set of smaller \textit{subregions}, corresponding to Voronoi cells of a set of \textit{subcentroids}. The naive solution of this problem via K-Means clustering in each region would require storing full-dimensional subcentroids codebooks that would require too much memory. Instead, we propose an almost memory-free approach that constructs the subcentroids codebook in each region as a set of convex combinations of the region centroid and its neighboring centroids. We refer to the proposed technique as \textit{grouping} procedure and describe it formally below.


\begin{figure*}
\begin{tabular}{cc}
\includegraphics[width=6cm]{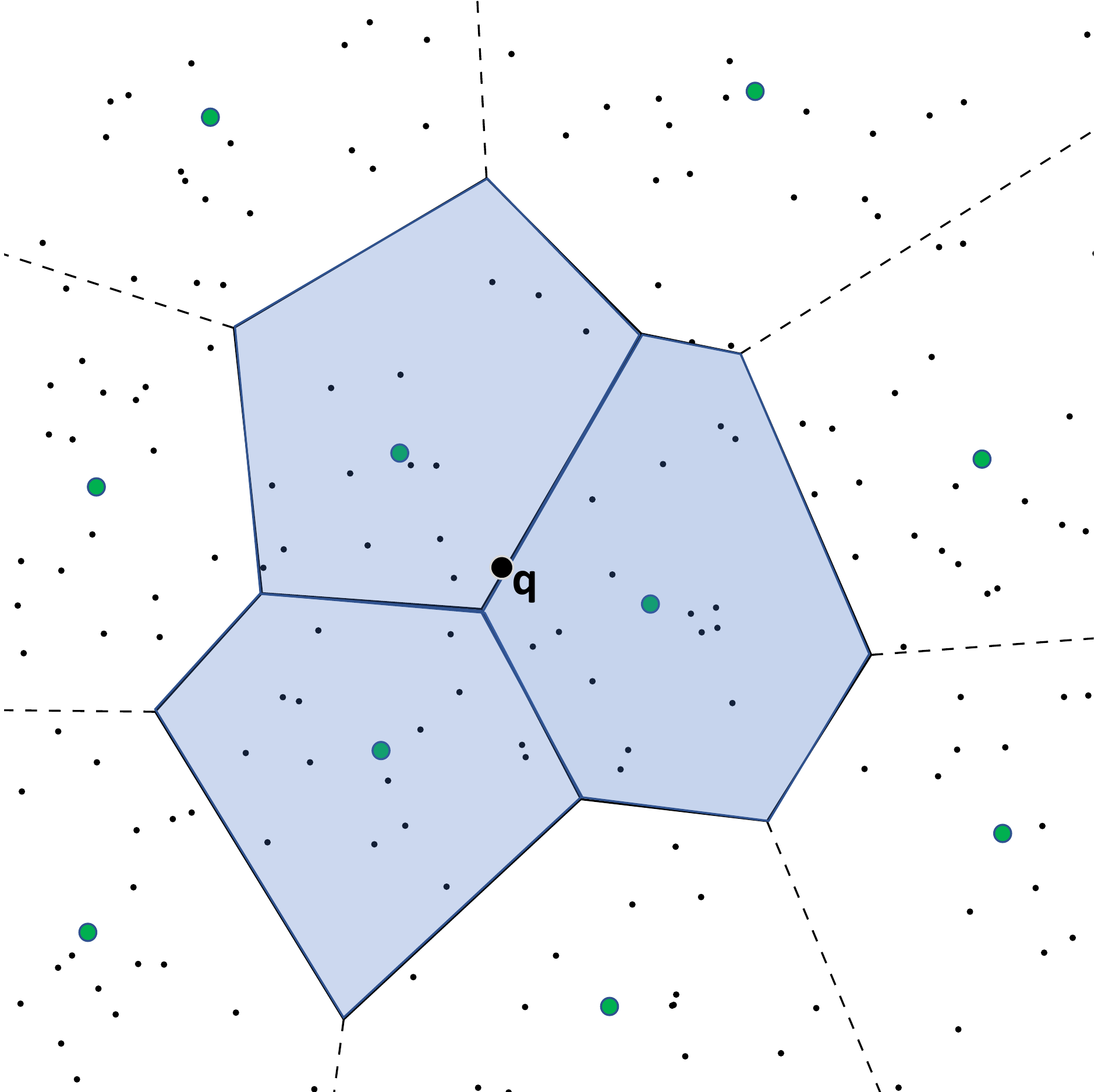}&
\includegraphics[width=6cm]{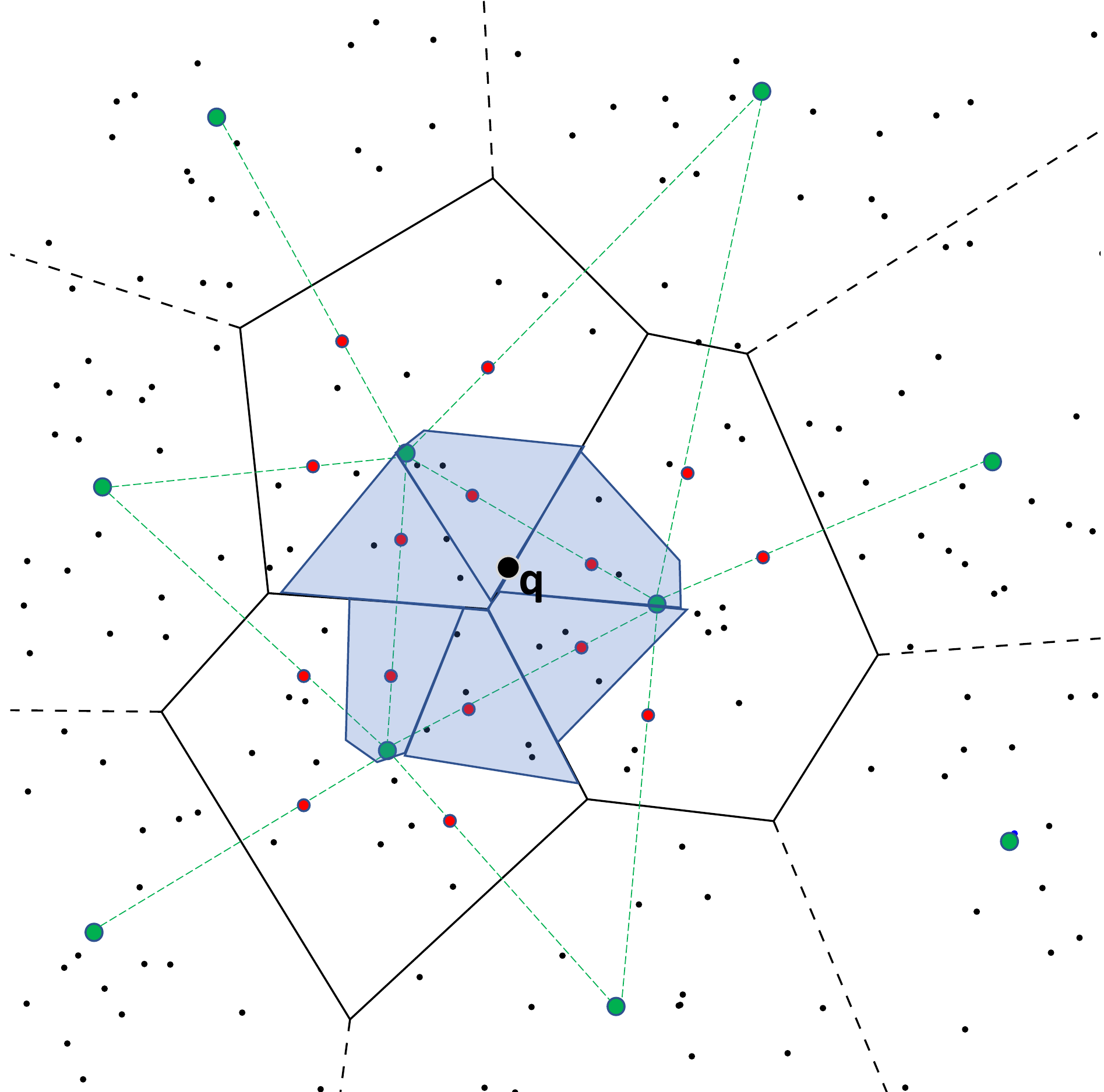}
\end{tabular}
\caption{The indexing and the search process for the dataset of $200$ two-dimensional points (small black dots) with the inverted index (left) and the inverted index augmented with grouping and pruning procedures (right). The large green points denote the region centroids, and for each centroid $L{=}5$ neighboring centroids are precomputed. For three regions in the center of the right plot, the region subcentroids are denoted by the red points. The fractions of the database traversed by the same query $q$ with and without pruning are highlighted in blue. Here the query is set to visit only $\tau{=}40\%$ closest subregions.}
\label{fig:teaser}
\end{figure*}


\textbf{The model.} The grouping procedure is performed independently for all the regions so it is sufficient to describe it for the single region with the centroid $c$. We assume that the database points $\{x_1,\dots,x_n\}$ belong to this region. Let us denote by $s_1,\dots,s_L \in C$ the nearest centroids of the centroid $c$:
\begin{gather}
\{s_1,\dots,s_L\} = \text{NN}_L(c)
\end{gather}
where $NN_L(c)$ denotes the set of $L$ nearest neighbors for $c$ in the set of all centroids.
The region subcentroids then taken to be $\{c + \alpha(s_l - c)\}, \ l=1,\dots,L$, where $\alpha$ is a scalar parameter that is learnt from data as we describe below. Note that different $\alpha$ values are used in different regions.
The points $\{x_1,\dots,x_n\}$ are distributed over Voronoi subregions produced by this set of subcentroids. For each point $x_i$ we determine the closest subcentroid 
\begin{gather}
l_i = \arg\min_{l}\|x_i - (c + \alpha(s_l - c))\|^2
\end{gather}
In the indexing structure the region points are stored in groups, i.e. all points from the same subregion are ordered continuously. In this scheme, we store only the subregion sizes to determine what group the particular point belongs to. After grouping, the displacements from the corresponding subcentroids
\begin{gather}
x_i - (c + \alpha(s_{l_i} - c))
\end{gather}
are compressed with PQ, as in the original IVFADC. Note that the displacements to subcentroids typically have smaller norms than the displacements to the region centroid as in the IVFADC scheme. Hence they could be compressed more accurately with the same code length. This results in higher recall rates of the retrieval scheme as will be shown in the experimental section.

\textbf{Distance estimation.} Now we describe how to compute the distances to the compressed points after grouping. One has to calculate an expression:
\begin{gather}
\|q - c - \alpha(s - c) - [r_1,\dots,r_M]\|^2
\label{eq:ivfadcdist_group}
\end{gather}
where the $[r_1,\dots,r_M]$ is the PQ approximation of the database point displacement.
The expression \eq{ivfadcdist_group} can be transformed in the following way:
\begin{gather}
\|q - c - \alpha(s - c) - [r_1,\dots,r_M]\|^2 = (1 - \alpha)\|q - c\|^2 + \nonumber \\
 + \alpha\|q - s\|^2 - 2\sum_{m=1}^{M}\langle q_m, r_m\rangle + const(q)
\label{eq:ivfadcdist_group_eff}
\end{gather}

The first term in the sum above can be easily computed as the distance $\|q - c\|^2$ is known from the closest centroids search result. The distances $\|q - s\|^2$ are computed online before visiting the region points. Note that the sets of neighboring centroids for the close regions typically have large intersections, and we do not recalculate the distances $\|q - s\|^2$, which were computed earlier for previous regions, for efficiency. The scalar products between the query subvectors and PQ codewords $\langle q_m, r_m\rangle$ are precomputed before regions traversal. The last term is query-independent, and we quantize it into $256$ values and explicitly keep its quantized value as an additional byte in the point code. Note that the computation of distances to the neighboring centroids results in additional runtime costs. In the experiments below we show that these costs are completely justified by the improvement in the compression accuracy. The number of subregions $L$ is set in such a way that the additional memory consumption ($K\cdot L\cdot ${\tt sizeof(float)} bytes) is negligible compared to the compressed database size.

\textbf{Subregions pruning.} The use of the grouping technique described above also allows the search procedure to skip the least promising subregions during region traversal. This provides the total search speedup without loss in search accuracy. Below we refer to such subregions skipping as \textit{pruning}. Let us describe pruning in more details. Consider traversing the particular region with a centroid $c$, the neighboring centroids $s_1,\dots,s_L$ and the scaling factor $\alpha$. The distances to the subcentroids can then be easily precomputed as follows:
\begin{gather}
\|q - c - \alpha(s_l - c)\|^2 = (1 - \alpha)\|q - c\|^2
 + \alpha\|q - s_l\|^2 + \text{const}(q),\ l=1\dots L
\end{gather}
In the sum above the first and the second terms are computed as described in the previous paragraph while the last term is precomputed offline and stored explicitly for each neighboring centroid. If the search process is set to visit $k$ inverted index regions, then $kL$ distances to the subcentroids are calculated, and only a certain fraction  $\tau$ of the closest subregions is visited. In practice, we observed that the search process could filter out up to half of the subregions without accuracy loss that provides additional search acceleration. \fig{teaser} schematically demonstrates the retrieval stage with and without pruning for the same query.

\textbf{Learning the scaling factor $\alpha$.} Finally, we describe how to learn the scaling factor $\alpha$ for the particular region with a centroid $c$ and the neighboring centroids $s_1,\dots,s_L$. 
$\alpha$ is learnt on the hold-out learning set, and we assume that the region contains the learning points $x_1,\dots,x_n$. We aim to solve the following minimization problem:
\begin{gather}
\min_{\alpha \in [0;1]}\sum_{i=1}^{n}{\min_{l_i}\|x_i - c - \alpha(s_{l_i} - c)\|^2}
\label{eq:alphalearn}
\end{gather}
In other words, we want to minimize the distances between the data points and the scaled subcentroids given that each point is assigned to the closest subcentroid. We also restrict $\alpha$ to belong to the $[0;1]$ segment so that each subcentroid is a convex combination of $c$ and one of the neighboring centroid $s$.

The exact solution of the problem above requires joint optimization over the continuous variable $\alpha$ and the discrete variables $l_i$. Instead, we solve \eq{alphalearn} approximately in two steps:
\begin{enumerate}
    \item First, for each training point $x_i$ we determine the optimal $s_{l_i}$ value. This is performed by minimizing the auxiliary function that is the lower bound of the target function in \eq{alphalearn}:
    \begin{gather}
    \sum_{i=1}^{n}{\min_{l_i,\alpha_{i} \in [0;1]}\|x_i - c - \alpha_{i}(s_{l_i} - c)\|^2}
  \label{eq:alphalearn_xi}
  \end{gather}
  This problem is decomposable into $n$ identical minimization subproblems for each learning point $x_i$:
  \begin{gather}
\min_{\alpha_i\in [0;1], s_{l_i}}\|x_i - c - \alpha_i(s_{l_i} - c)\|^2
\label{eq:alphalearn_subproblem}
\end{gather}
This subproblem is solved via exhaustive search over all possible $s_{l_i}$. For a fixed $s_{l_i}$, the minimization over $\alpha_i$ has a closed form solution and the corresponding minimum value of the target function \eq{alphalearn_subproblem} can be explicitly computed. Then the solution of the subproblem \eq{alphalearn_subproblem} for the point $x_i$ is:
\begin{gather}
s_{l_i}^{*} = \arg\min_{s_{l_i}}\left|\left|x_i - c - \frac{(x_i - c)^{T}(s_{l_i} - c)}{\|s_{l_i} - c\|^2}(s_{l_i} - c)\right|\right|^2
\end{gather}
    \item Second, we minimize \eq{alphalearn} over $\alpha$ with the values of $s_{l_i}^{*}$ obtained from the previous step. In this case the closed-form solution for the optimal value is:
    \begin{gather}
    \alpha = \frac{\sum_{i=1}^{n}(x_i - c)^T(s_{l_i}^{*} - c)}{\sum_{i=1}^{n}\|s_{l_i}^{*} - c\|^2}
    \end{gather}
\end{enumerate}

\textbf{Discussion.} The grouping and pruning procedures described above allow to increase the compression accuracy and the candidate lists quality. This results in a significant enhancement in the final system performance as will be shown in the experimental section. Note that these procedures are more effective for the inverted index, and they cannot be exploited as efficiently in the IMI due to a very large number of regions in its space partition.

\section{Experiments}
In this section we present the experimental comparison of the proposed indexing structure and the corresponding retrieval system with the current state-of-the-art.

\textbf{Datasets.} We perform all the experiments on the publicly available datasets that are commonly used for billion-scale ANN search evaluation:
\begin{enumerate}
    \item DEEP1B dataset\cite{BabenkoCVPR16} contains one billion of $96$-dimensional CNN-produced feature vectors of the natural images from the Web. The dataset also contains a learning set of 350 million descriptors and 10,000 queries with the groundtruth nearest neighbors for evaluation.
    \item SIFT1B dataset\cite{Jegou11b} contains one billion of $128$-dimensional SIFT descriptors as a base set, a hold-out learning set of 100 million vectors, and 10,000 query vectors with the precomputed groundtruth nearest neighbors.
\end{enumerate}
 
In most of the experiments the search accuracy is evaluated by the $Recall@R$ measure which is calculated as a rate of queries for which the true nearest neighbor is presented in the short-list of length $R$. All trainable parameters are obtained on the hold-out learning sets. All experiments are performed on the Intel Xeon E5-2650 2.6GHz CPU in a single thread mode.

\textbf{Large codebooks in the inverted index.} As we show in \sect{methods} the indexing quality of the inverted index does not saturate even with codebooks of several million centroids. As the exhaustive query assignment would be inefficient for large codebooks, we use the approximate nearest centroids search via HNSW algorithm\cite{HNSW}. The algorithm is based on the proximity graph, constructed on the set of centroids. As we observed in our experiments, HNSW allows obtaining a small top of the closest centroids with almost perfect accuracy in a submillisecond time. We also use HNSW on the codebooks learning stage to accelerate the assignment step during K-Means iterations. The only cost of the HNSW search is the additional memory required to maintain the proximity graph. In our experiments each vertex of the proximity graph is connected to $32$ other vertices, hence the total memory for all the edge lists equals $32\cdot K\cdot${\tt sizeof(int)} bytes, where $K$ denotes the codebook size. 

Note that the accuracy and efficiency of the HNSW are crucial for the successful usage of large codebooks with an approximate assignment. The earlier efforts to use larger codebooks were not successful: \cite{Babenko12} evaluated the scheme based on the inverted index with a very large codebook where the closest centroids were found via kd-tree\cite{Bentley75}. It was shown that this scheme was not able to achieve the state-of-the-art recall rates due to inaccuracies of the closest centroids search. 

\begin{figure}
\noindent
\centering
\begin{tabular}{cc}
DEEP1B & 
SIFT1B \\
\includegraphics[height=4.8cm]{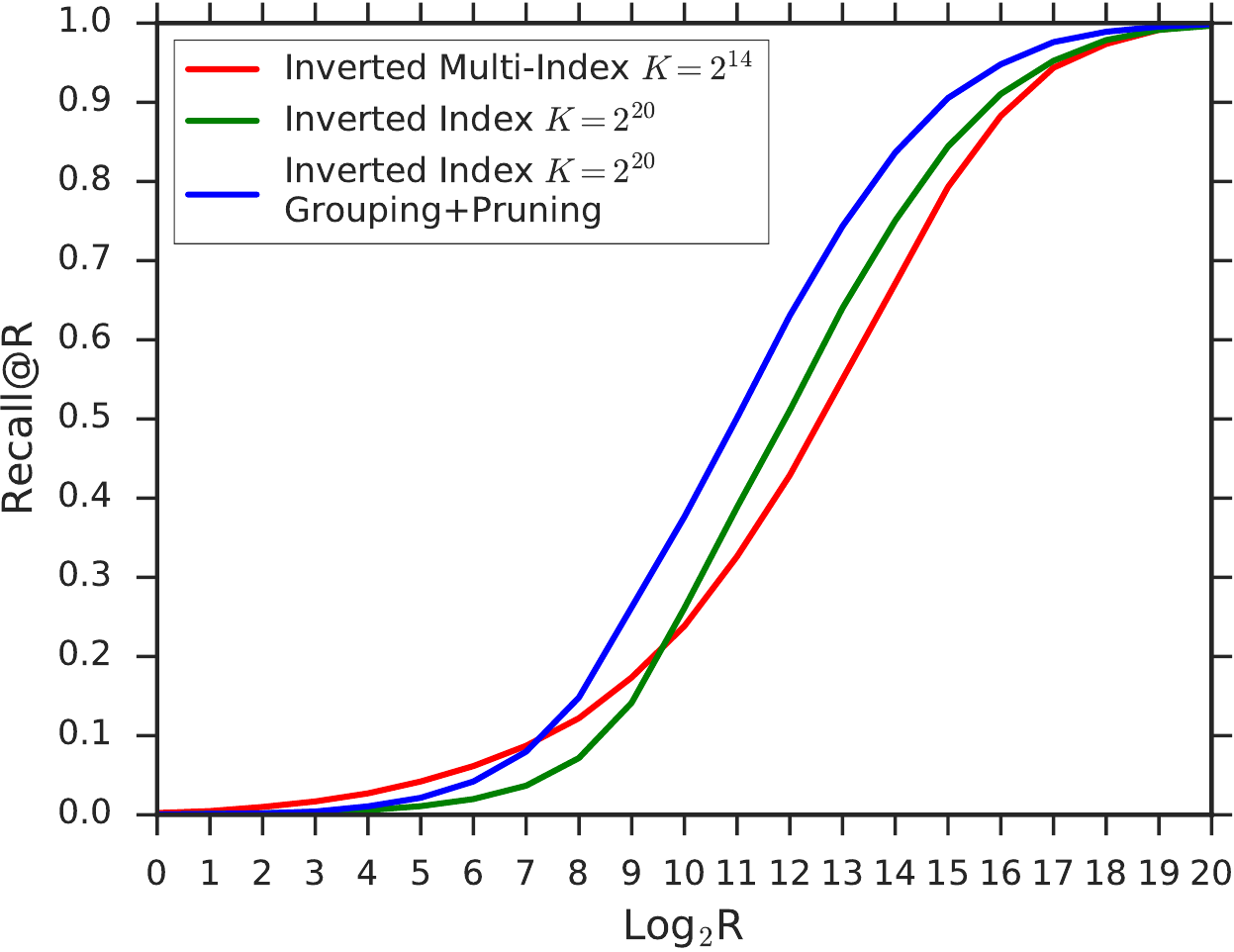}
& 
\includegraphics[height=4.8cm]{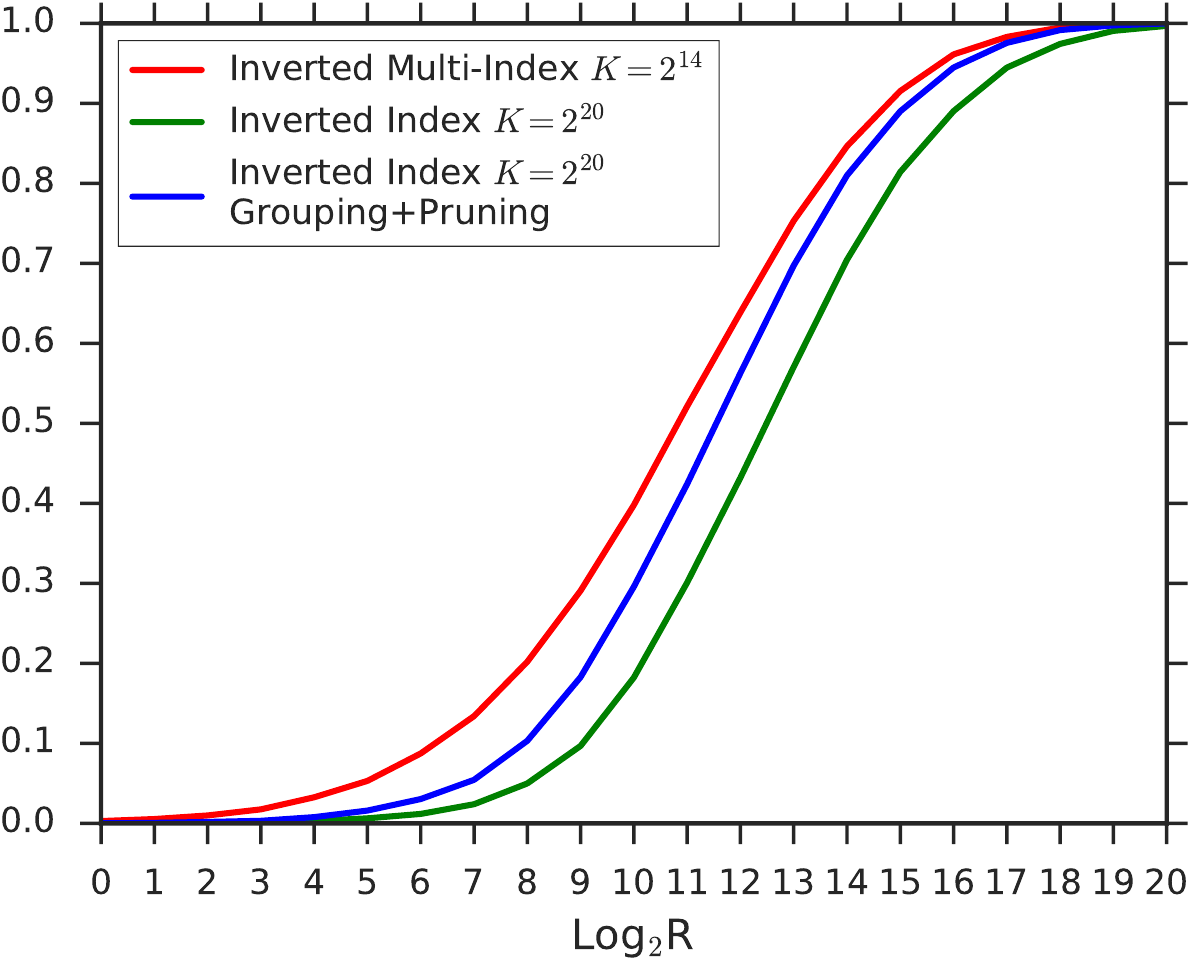}\\
\end{tabular}
\caption{Recall as a function of the candidate list length for inverted multi-indices with $K{=}2^{14}$, inverted index with $K{=}2^{20}$ with and without pruning. On DEEP1B the inverted indices outperform the IMI for all reasonable values of $R$ by a large margin. For SIFT1B the candidate lists quality of the inverted index with pruning is comparable to the quality of the IMI for $R$ larger than $2^{13}$.} 
\label{fig:shortlist}
\end{figure}

\textbf{Indexing quality.} In the first experiment we evaluate the ability of different indexing approaches to extract concise and accurate candidate lists. The candidates reranking is not performed here. We compare the following structures:

\begin{enumerate}
\item \textbf{Inverted Multi-Index (IMI)} \cite{Babenko12}. We evaluate the IMI with codebooks of size $K=2^{14}$ and consider the variant of the IMI with global rotation before dataspace decomposition~\cite{OpqTr} that boosts the IMI performance on datasets of deep descriptors. In all experiments we used the implementation from the FAISS library\cite{FAISS}.

\item \textbf{Inverted Index}\cite{Sivic03}. We use a large codebook of $K{=}2^{20}$ centroids. The query assignment is performed via HNSW.

\item \textbf{Inverted Index + Grouping + Pruning}. Here we augment the inverted index setup from above with the grouping and pruning procedures described in \sect{grouping}. The number of subregions is set to $L{=}64$, and the pruning ratio is set to $\tau{=}50\%$.
\end{enumerate}

The $Recall@R$ values for different values of $R$ are demonstrated in \fig{shortlist}. Despite a much smaller number of regions, the inverted index produces more accurate short-lists compared to the IMI for the DEEP1B dataset. Note that the pruning procedure in the inverted index improves short-lists quality even further. The most practically important part of this plot corresponds to $R=10^{4}-10^{5}$ and in this range the inverted index outperforms the IMI by up to $10\%$.

For the SIFT1B dataset, the IMI with $K{=}2^{14}$ produces a slightly better candidate lists for small values of $R$. For $R>2^{13}$ the quality of the inverted index is comparable to the IMI quality. The IMI is successful on SIFT vectors, as they are histogram-based and the subvectors corresponding to the different halves of them describe disjoint image parts that typically have relatively weak statistical inter-dependency. However, as we show in the next experiment, the runtime cost of candidates extraction in the IMI is high due to the inefficiency of the multi-sequence algorithm and a large number of random memory accesses.

\textbf{ANN: indexing + reranking.} As the most important experiment, we evaluate the performance of the retrieval systems built on top of the aforementioned indexing structures for approximately the same memory consumption. All the systems operate in the compressed domain, i.e.\ the displacements of database points from their region centroids are OPQ-compressed with code lengths equal to $8$ or $16$ bytes per point.
In this experiment candidate lists are reranked based on the distances between the query and the compressed candidate points. The OPQ codebooks are global and shared by all regions. We compare the following systems:
\begin{figure*}
\noindent
\centering
\begin{tabular}{cc}
\includegraphics[height=4.6cm]{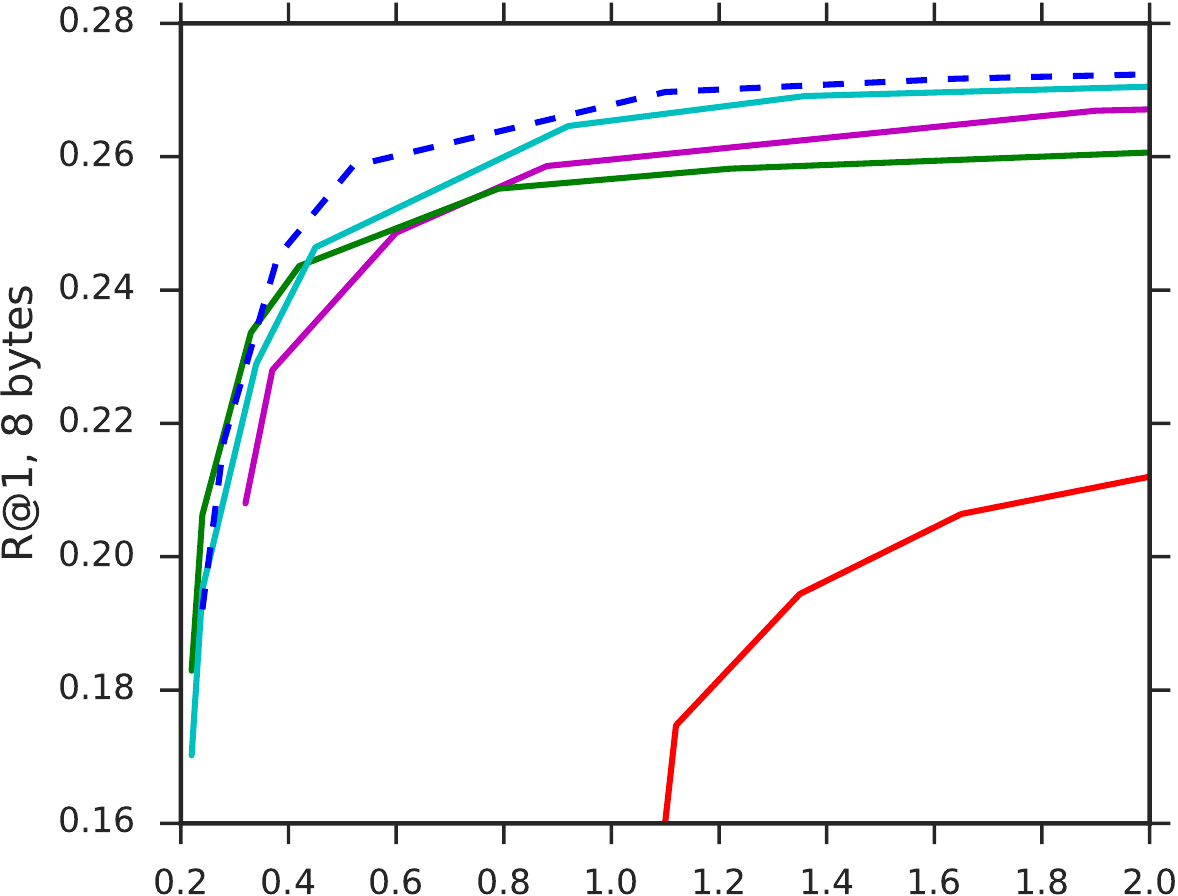}&
\includegraphics[height=4.6cm]{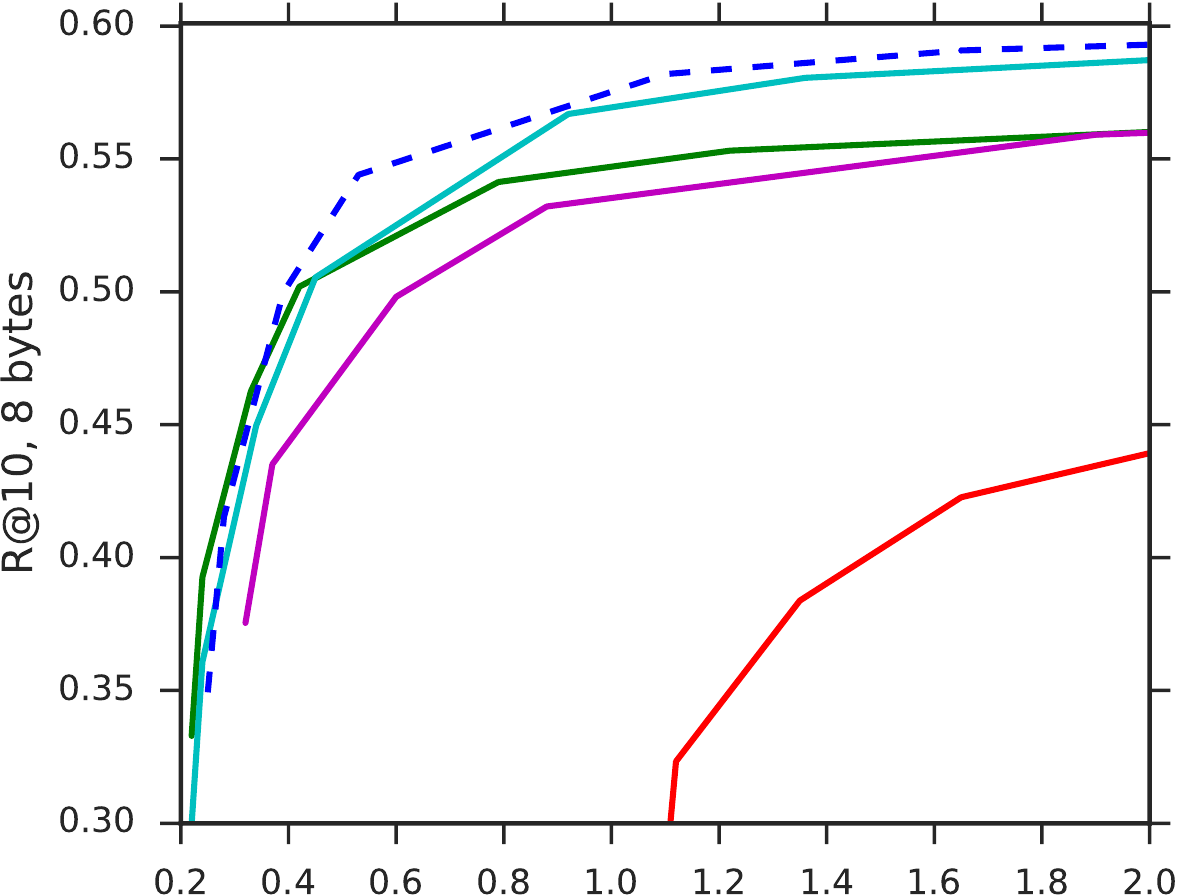}\\
\includegraphics[height=4.9cm]{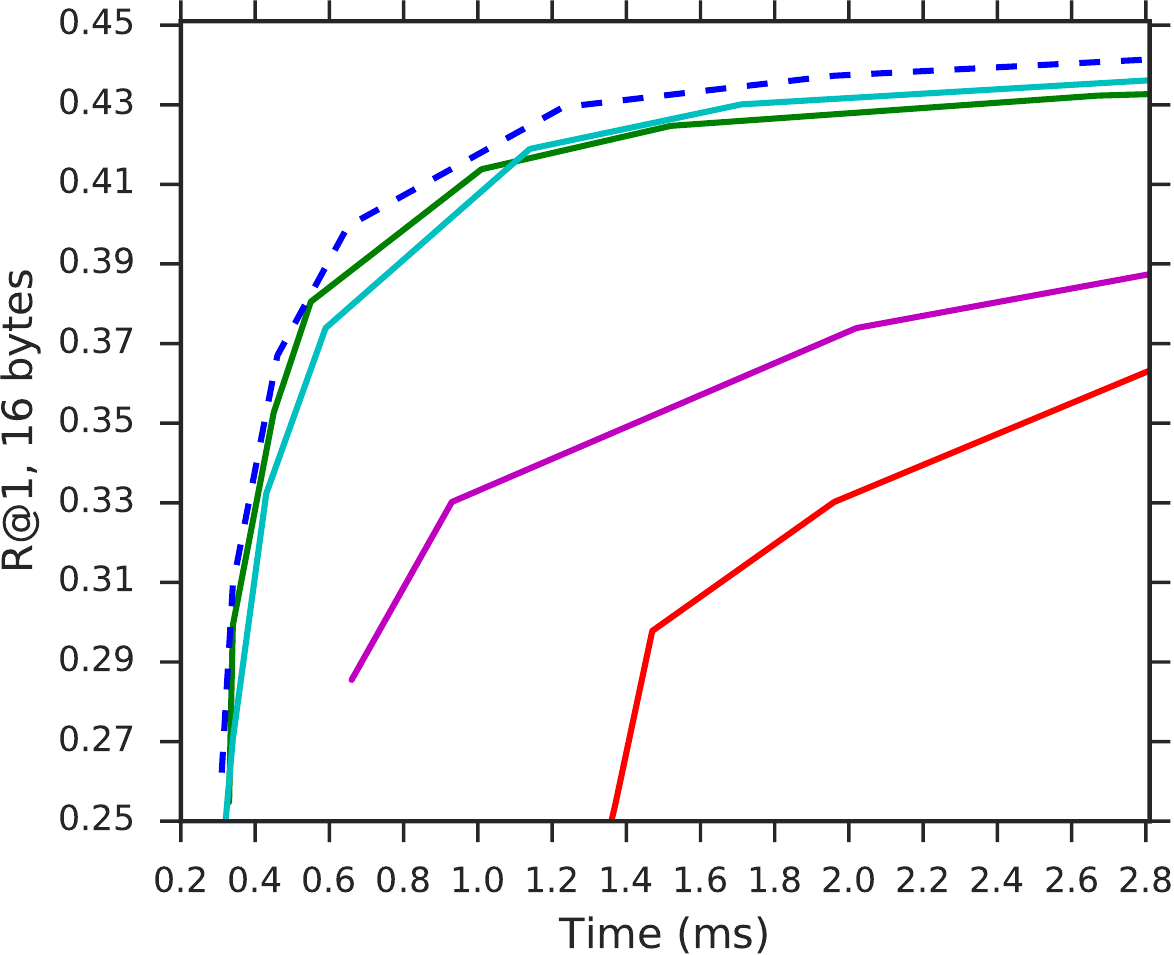}&
\includegraphics[height=4.9cm]{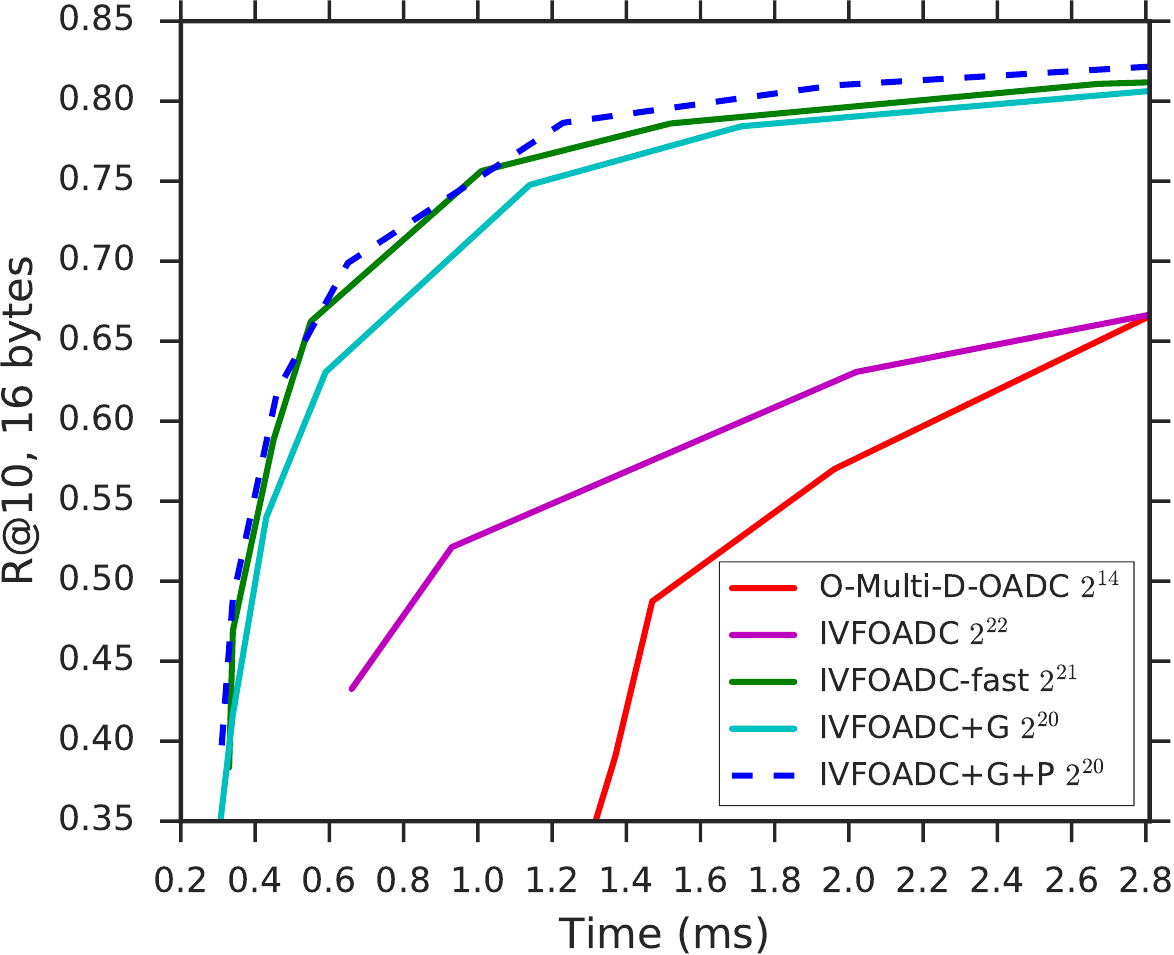}
\end{tabular}
\caption{The $R@1$ and $R@10$ values after reranking as functions of runtime on the DEEP1B. The systems based on the inverted index substantially outperform the IMI-based system. The IVFOADC system with grouping outperforms the IVFOADC systems with larger codebooks for the same memory consumption.} 
\label{fig:rerank}
\end{figure*}
\begin{enumerate}
\item \textbf{O-Multi-D-OADC} is our main baseline system. It uses the inverted multi-index with global rotation and a codebook of size $K{=}2^{14}$. This system requires 1 Gb of additional memory to maintain the IMI structure.
\item \textbf{IVFOADC} is based on the inverted index with a codebook of a size $K{=}2^{22}$. This system requires $2.5$Gb of additional memory to store the codebook and the HNSW graph.
\item \textbf{IVFOADC-fast} is a system that uses the expression \eq{ivfadcdist_group_eff} for efficient distance estimation with $\alpha=0$. This system is also based on the inverted index without grouping but requires one additional code byte per point to store the query-independent term from \eq{ivfadcdist_group_eff}.
We use $K{=}2^{21}$ for this scheme to make the total memory consumption the same as for the previous system. The memory consumption includes $1$Gb for the additional code bytes and $1.25$Gb to store the codebook and the graph that gives $2.25$Gb in total.

\item \textbf{IVFOADC+Grouping} additionally employs the grouping procedure with $L{=}64$ subcentroids per region. In this system we use a codebook with $K{=}2^{20}$ that results in the total memory consumption of $1.87$Gb. 
\item \textbf{IVFOADC+Grouping+Pruning} employs both grouping and pruning procedures with $L{=}64$ subcentroids. The pruning is set to filter out $50\%$ of the subregions. In this system we also use a codebook with $K{=}2^{20}$.
\end{enumerate}

We plot $Recall@1$ and $Recall@10$ on the DEEP1B dataset for different lengths of candidate lists as functions of the corresponding search runtime. The results are summarized in \fig{rerank}. We highlight several key observations:

\begin{enumerate}
    \item The systems based on the inverted index outperform the IMI-based system in terms of accuracy and search time. In particular, for a time budget of 1.5 ms, the IVFOADC+G+P system outperforms the O-Multi-D-OADC by $7$ and $17$ percent points of $R@1$ and $R@10$ respectively on the DEEP1B dataset and 8-byte codes. As for the runtime, this system reaches the same recall values several times faster compared to O-Multi-D-OADC.
    \item The IVFOADC system with grouping and pruning outperforms the IVFOADC systems with larger codebooks without grouping. The advantage is the most noticeable for short 8-byte codes when an additional encoding capacity from grouping is more valuable.
\end{enumerate}

\textbf{The inverted multi-index limitations.} Here we perform several experiments to demonstrate that both approximate query assignment and grouping are more beneficial for IVFADC than for IMI. In theory, one could also accelerate the IMI-based schemes via using approximate closest subspace centroids search. However, in this case, one would have to find several hundred closest items from a moderate codebook of size $K{=}2^{14}$, and we observed that in this setup the approximate search with HNSW takes almost the same time as brute-force. Moreover, such acceleration would not speed up the candidates accumulation that is quite slow in the multi-index due to a large number of empty regions.

Second, the grouping procedure is less effective for the IMI compared to the inverted index. With $K{=}2^{14}$ each region in the IMI space partition contains only a few points, hence grouping is useless. To evaluate grouping effectiveness for the IMI with coarser codebooks we perform the following experiment. We compute the relative decrease in the average distance from the datapoints to the closest (sub-)centroid before and after grouping with $L{=}64$. Here we compare the inverted index with $K{=}2^{20}$ and the IMI with $K{=}2^{10}$ that result in the space partitions with the same number of regions. The average distances before and after grouping are presented in \tab{ablation}, right. The relative decrease in the average distances is smaller for the IMI that implies that grouping is more effective for the inverted index compared to the IMI. However, we assume that one of the interesting research directions is to investigate if the grouping could be incorporated in the IMI effectively.

\begin{table}
\begin{center}
\begin{tabular}{c c}
\begin{tabular}{|c|ccc|c|}
\hline
$L$ & R@1 & R@10& R@100 & $t(ms)$\\
\hline
32 & 0.417 & 0.776 & 0.869 & 1.22\\
\hline
64 & 0.433 & 0.785 & 0.878 & 1.28 \\
\hline 
128 & 0.441 & 0.791 & 0.882 & 1.48\\
\hline
\end{tabular}
&
\begin{tabular}{|c|c|c|}
\hline
 &
\multicolumn{1}{|c|}{Inverted Index} & 
\multicolumn{1}{|c|}{Inverted Multi-Index}\\
\hline
No grouping & 0.282 & 0.415\\
\hline
With grouping & 0.255 & 0.385\\
\hline
Decrease & \textbf{10\%} & 7\%\\
\hline
\end{tabular}
\end{tabular}
\end{center}
\caption{$Left$: The recall values and the runtimes of the IVFOADC+Grouping+Pruning system for different numbers of subcentroids per region on the DEEP1B dataset. Here we use the candidate lists of length $30K$ and 16-byte codes. $Right$: The average distances from the datapoints to the closest (sub-)centroids with and without grouping for the inverted index with $K=2^{20}$ and the IMI with $K=2^{10}$ on the DEEP1B dataset.} 
\label{tab:ablation}
\end{table}

\textbf{Number of grouping subregions.} We also demonstrate the performance of the proposed scheme for different numbers of subcentroids per region $L$. In \tab{ablation}, left we provide the evaluation of the IVFOADC+Grouping+Pruning system on DEEP1B for candidate lists of size $30K$ and 16-byte codes. The usage of $L>64$ is hardly justified due to increase in runtime and memory consumption.

\textbf{Comparison to the state-of-the-art.} Finally, we compare the proposed IVFADC+G+P with the results reported in the literature on the DEEP1B and SIFT1B, see \tab{soa}. Along with the recall values and timings we also report the amount of additional memory per point, required by each system.

\begin{table}
\small
\centering
\renewcommand\arraystretch{0.9}
\begin{tabular}{|c|c|c|c|c|c|c|c|c|c|c|c|}
\hline
\multicolumn{2}{|c|}{} & \multicolumn{5}{|c|}{DEEP1B} & \multicolumn{5}{|c|}{SIFT1B}\\
\hline
\small{Method} & $K$ & \small{R@1} & \small{R@10} & \small{R@100} & t & \small{Mem} & \small{R@1} & \small{R@10} & \small{R@100} & t & \small{Mem}\\
\hline
O-Multi-D-OADC\cite{BabenkoTPAMI15} & $2^{14}$ & 0.397 & 0.766 & 0.909 & 8.5 & 17.34 & 0.360 & 0.792 & 0.901 & 5 & 17.34\\
\hline
Multi-LOPQ\cite{Kalantidis14} & $2^{14}$ & 0.41 & 0.79 & - & 20 & 18.68 & \textbf{0.454} & \textbf{0.862} & 0.908 & 19 & 19.22\\
\hline
GNOIMI\cite{BabenkoCVPR16} & $2^{14}$  & 0.45 & 0.81 & - & 20 & 19.75 & - & - & - & - & -\\
\hline
IVFOADC+G+P & $2^{20}$  & \textbf{0.452} & \textbf{0.832} & \textbf{0.947} & \textbf{3.3} & 17.87 & 0.405 & 0.851 & \textbf{0.957} & \textbf{3.5} & 18\\
\hline
\end{tabular}

\caption{Comparison to the previous works for 16-byte codes. The search runtimes are reported in milliseconds. We also provide the memory per point required by the retrieval systems (the numbers are in bytes and do not include 4 bytes for point ids).}
\label{tab:soa}
\end{table}

\section{Conclusion}

In this work, we have proposed and evaluated a new system for billion-scale nearest neighbor search. The system expands the well-known inverted index structure and makes no assumption about database points distribution what makes it a universal tool for datasets with any data statistics. The advantage of the scheme is demonstrated on two billion-scale publicly available datasets.

\bibliographystyle{splncs}
\bibliography{egbib}

\begin{thebibliography}{10}

\bibitem{Jegou11b}
Jegou, H., Tavenard, R., Douze, M., Amsaleg, L.:
\newblock Searching in one billion vectors: Re-rank with source coding.
\newblock In: ICASSP. (2011)

\bibitem{Babenko12}
Babenko, A., Lempitsky, V.S.:
\newblock The inverted multi-index.
\newblock In: 2012 {IEEE} Conference on Computer Vision and Pattern
  Recognition, Providence, RI, USA, June 16-21, 2012. (2012)

\bibitem{OpqTr}
Ge, T., He, K., Ke, Q., Sun, J.:
\newblock Optimized product quantization.
\newblock Technical report (2013)

\bibitem{Kalantidis14}
Kalantidis, Y., Avrithis, Y.:
\newblock Locally optimized product quantization for approximate nearest
  neighbor search.
\newblock In: in Proceedings of International Conference on Computer Vision and
  Pattern Recognition (CVPR 2014), IEEE (2014)

\bibitem{BabenkoCVPR16}
Babenko, A., Lempitsky, V.S.:
\newblock Efficient indexing of billion-scale datasets of deep descriptors.
\newblock In: CVPR. (2016)

\bibitem{FAISS}
Johnson, J., Douze, M., J{\'e}gou, H.:
\newblock Billion-scale similarity search with gpus.
\newblock arXiv preprint arXiv:1702.08734 (2017)

\bibitem{Philbin07}
Philbin, J., Chum, O., Isard, M., Sivic, J., Zisserman, A.:
\newblock Object retrieval with large vocabularies and fast spatial matching.
\newblock In: CVPR. (2007)

\bibitem{Douze_2018_CVPR}
Douze, M., Szlam, A., Hariharan, B., Jegou, H.:
\newblock Low-shot learning with large-scale diffusion.
\newblock In: CVPR. (2018)

\bibitem{Wang_2017_TPAMI}
Wang, D., Otto, C., Jain, A.K.:
\newblock Face search at scale.
\newblock TPAMI (2017)

\bibitem{Jegou11a}
J{\'e}gou, H., Douze, M., Schmid, C.:
\newblock Product quantization for nearest neighbor search.
\newblock TPAMI \textbf{33}(1) (2011)

\bibitem{Ge13}
Ge, T., He, K., Ke, Q., Sun, J.:
\newblock Optimized product quantization for approximate nearest neighbor
  search.
\newblock In: CVPR. (2013)

\bibitem{Norouzi13}
Norouzi, M., Fleet, D.J.:
\newblock Cartesian k-means.
\newblock In: CVPR. (2013)

\bibitem{BabenkoCVPR14}
Babenko, A., Lempitsky, V.:
\newblock Additive quantization for extreme vector compression.
\newblock In: CVPR. (2014)

\bibitem{BabenkoCVPR15}
Babenko, A., Lempitsky, V.S.:
\newblock Tree quantization for large-scale similarity search and
  classification.
\newblock In: CVPR. (2015)

\bibitem{CQ}
Zhang, T., Du, C., Wang, J.:
\newblock Composite quantization for approximate nearest neighbor search.
\newblock In: ICML. (2014)

\bibitem{SCQ}
Zhang, T., Qi, G.J., Tang, J., Wang, J.:
\newblock Sparse composite quantization.
\newblock In: CVPR. (2015)

\bibitem{MartinezECCV16}
Martinez, J., Clement, J., Hoos, H.H., Little, J.J.:
\newblock Revisiting additive quantization.
\newblock In: ECCV. (2016)

\bibitem{Polysemous}
Douze, M., J{\'{e}}gou, H., Perronnin, F.:
\newblock Polysemous codes.
\newblock In: ECCV. (2016)

\bibitem{JegouECCV16}
Jain, H., P{\'{e}}rez, P., Gribonval, R., Zepeda, J., J{\'{e}}gou, H.:
\newblock Approximate search with quantized sparse representations.
\newblock In: ECCV. (2016)

\bibitem{Sivic03}
Sivic, J., Zisserman, A.:
\newblock Video google: {A} text retrieval approach to object matching in
  videos.
\newblock In: ICCV. (2003)

\bibitem{Wieschollek_2016_CVPR}
Wieschollek, P., Wang, O., Sorkine-Hornung, A., Lensch, H.P.A.:
\newblock Efficient large-scale approximate nearest neighbor search on the gpu.
\newblock In: CVPR. (2016)

\bibitem{HNSW}
Malkov, Y.A., Yashunin, D.A.:
\newblock Efficient and robust approximate nearest neighbor search using
  hierarchical navigable small world graphs.
\newblock arXiv preprint arXiv:1603.09320 (2016)

\bibitem{Bentley75}
Bentley, J.L.:
\newblock Multidimensional binary search trees used for associative searching.
\newblock Commun. ACM \textbf{18}(9) (1975)

\bibitem{BabenkoTPAMI15}
Babenko, A., Lempitsky, V.S.:
\newblock The inverted multi-index.
\newblock {IEEE} Trans. Pattern Anal. Mach. Intell. \textbf{37}(6) (2015)
  1247--1260

\end{thebibliography}

\end{document}